\documentclass[11pt,a4paper]{article}

\usepackage{amsmath}
\usepackage{amssymb}
\usepackage{amsthm}
\usepackage{color}
\usepackage{graphicx}
\usepackage{nicefrac}
\usepackage{sfmath}
\usepackage[left=2.5cm,right=2cm,%
           top=1.5cm,bottom=2cm,%
           includefoot,includehead]{geometry}
\usepackage{authblk}
\usepackage{algorithm}
\usepackage{algorithmicx}
\usepackage{algpseudocode}
\usepackage{units}

\allowdisplaybreaks{}

\author[1]{Keyan Ghazi-Zahedi \thanks{zahedi@mis.mpg.de}}
\affil[1]{Max Planck Institute for Mathematics in the Sciences, Leipzig, Germany}

\title{NMODE --- Neuro-MODule Evolution}
\begin{document}
\maketitle

\begin{abstract}
Modularisation, repetition, and symmetry are structural features shared by
almost all biological neural networks. These features are very unlikely to be
found by the means of structural evolution of artificial neural networks. This
paper introduces {\sf NMODE}, which is specifically designed to operate on
neuro-modules. {\sf NMODE} addresses a second problem in the context of
evolutionary robotics, which is incremental evolution of complex behaviours for
complex machines, by offering a way to interface neuro-modules. The scenario in
mind is a complex walking machine, for which a locomotion module is evolved
first, that is then extended by other modules in later stages. We show that {\sf
NMODE} is able to evolve a locomotion behaviour for a standard six-legged
walking machine in approximately 10 generations and show how it can be used for
incremental evolution of a complex walking machine. The entire source code used
in this paper is publicly available through GitHub.
\end{abstract}

\section{Introduction}
\label{sec:introduction}
One of the advantages of evolutionary algorithms is that these type of
algorithms can be used to find
novel solutions with a minimal amount of pre-structuring. That means that
there is no inherent need to e.g.~a priori specify the structure of a neural
network. In the context of embodied artificial intelligence, this means that for
an embodied agent, only the number of sensors and motors need to be specified
and everything else can be left open to evolution. In particular, artificial
evolution can be used to find optimal control structures by sequentially growing
and pruning networks. We use the term optimal in the sense
of~\cite{Montufar2015aA-Theory}, i.e.~in the context of structurally minimal
networks that solve a given set of tasks. Furthermore, evolutionary algorithms
can be used to co-evolve the brain and body of embodied agents. Impressive
examples date back to Karl Sim's
work~\cite{Sims1994aEvolving,Sims1994bEvolving}, but can also be found in more
recent
work~(e.g.~\cite{Auerbach2014aEnvironmental,Nicholas-Cheney2014aEvolved}).

A common trait of almost all experiments (early as well as contemporary) in the
context of evolutionary robotics are that the morphology or the behaviour
is kept simple, while the other might vary. In cases of simple morphologies,
i.e.~wheel-driven robots, one can find examples of interesting behaviours such
as predator-prey scenarios~\cite{Nolfi2000aEvolutionary}. If the morphology is
more complex, i.e.~walking machines, then evolved behaviours, mainly solve
locomotion problems~(e.g.~\cite{Twickel2012aNeural}), which does not mean that
the results are less impressive. Yet, it indicates that it is a challenge to
evolve complex behaviours for complex systems. We believe that these challenges
result from two main problems. In the early days, genetic
algorithms~\cite{Holland1992aAdaptation} (GAs) were used predominantly. GAs were
often applied to fully connected recurrent neural networks, which means that the
string length would increased exponentially with every additional sensor or
motor. Hence, for simple systems with limited sensors and motors, the space of
policies was small enough to search for interesting behaviours, whereas for
complex morphologies, such as walking machines, evolving a robust walking
behaviour already posed a challenge. A workaround was to use neural networks of
particular structures, but then the question arose, what are best suited
structures to learn a specific behaviour? An interesting alternative, which is
known as HyperNEAT, will be discussed below. The second problem, in our opinion,
is that it is very difficult to add functionality to a network in an
evolutionary setting. Let us assume that we have evolved a locomotion network
for a hexapod. How can we add the functionality to add e.g.~a light-seeking
behaviour, while ensuring that the already learned behaviour is preserved. One
can fixate the synaptic weights of the locomotion network, but how should the
newly grown structure interface with the already existing locomotion structure?
{\sf NMODE} is specifically designed to address these two problems, i.e., to
reduce the search space for the evolution of neuro-modules in a meaningful way
and to provide a principled way to interface new module ,,on-top'' of already
evolved structures. Before we introduce {\sf NMODE}, we first present
contemporary algorithms and discuss why we did not choose to extend an existing
framework instead of creating yet another artificial evolution algorithm.

To the best of the authors knowledge, there is only one algorithm that can be
considered anything close to a standard in this
context.
HyperNEAT~\cite{Risi2012aAn-enhanced} is a very popular evolutionary algorithm
that allows to simultaneously evolve morphologies and neural networks. To
understand how HyperNEAT works, it is important to look at its predecessors,
NEAT~\cite{Stanley2002aEvolving} and CPPN~\cite{Secretan2011aPicbreeder:}. NEAT
is a very interesting algorithm to evolve neural networks of arbitrary
structure. Two key features of NEAT are its ability to allow for cross-over
between to arbitrary structures and speciation as a method to protect
innovations. CPPNs (compositional pattern producing networks) are NEAT
networks which use a set of function instead of the standard sigmoidal transfer
function. This means that one parameter of the neuron selects the transfer
function from a set of possible functions (sine wave, linear, saw tooth, \ldots).
Let us assume, that we have evolved a CPPN with two inputs and a three outputs.
Let us further assume, that the two inputs are the $x$ and $y$ coordinates of a
two-dimensional bounded plane and the three outputs are RGB colour values. This means
that the CPPN is now able to produce a picture. The Picbreeder
website\footnote{\url{http://www.picbreeder.com}}~\cite{Secretan2011aPicbreeder:}
shows impressive examples of pictures that were evolved by visitors of the website. It
is striking how CPPNs can make use of symmetries, repetitions, etc., to create
complex and appealing images. The website Endless
Forms\footnote{\url{http://www.endlessforms.com}} applies the same idea to 3D
shapes~\cite{Clune2011aEvolving}. The output of the CPPN is now used to
determine the boundary of geometric figures.
HyperNEAT uses CPPNs to generate the synaptic weights in a layered neural
network. Each layer of a neural network is placed on a two-dimensional plane
and two layers are fully connected, which means that every neuron in the input
layer is connected with every neuron in the hidden layer and every neuron in
the hidden layer is connected to every neuron in the output layer. A CPPN is
now fed with the geometric location of the pre- and post-synaptic neurons on
their layers and the output of the CPPN is the synaptic connection between the
two neurons. Although there are very interesting experiments published that
use HyperNEAT~(e.g.~\cite{Auerbach2014aEnvironmental}), there are currently no
publications that show a successful application of HyperNEAT to complex robotic
systems. For example, HyperNEAT has not been used to evolve behaviours for
non-trivial walking machines such as {\sf Svenja} or even {\sf Heaxaboard},
which is discussed in Section~\ref{sec:hexaboard} and
Section~\ref{sec:svenja}. One reason is that HyperNEAT is very good at finding
patterns, but not very good in fine tuning
parameters~\cite{Clune2009aEvolving}.

From our perspective, the significant reason not to use HyperNEAT is the its
requirement to use a predefined neural network architecture. The number of
layers and their respective connectivity must be specified by the experimenter
before running the evolutionary algorithm. We are interested in finding minimal
and optimal neural network architectures, which is why the structure of the
network (including the number of synapses and neurons) must be be open to
evolution. NEAT allows to evolve the structure of a network together with its
parameters but does not allow for modularisation. One idea that will be followed
in future work is to combine NEAT with the modular structure of {\sf NMODE}.
This would allow to use cross-over operators not only on the level of modules
(see Sec.~\ref{sec:nmode}) but also allow cross-over between modules.

Besides NEAT, there aren't many algorithms that support the co-evolution of
structure and parameters of a neural network. One of the first frameworks
probably is ISEE (integrated structure evolution environment), which is
described in detail in~\cite{Ghazi-Zahedi2009aSelf-Regulating} and based on the
$ENS^3$ algorithm~\cite{Dieckmann1995bCoevolution}. ISEE not only included
software to evolve neural networks but also sophisticated tools to inspect the
dynamics of neural network while they were operating in the sensorimotor loop
as wells as {\em ex vitro\/} analysis of their dynamics. It has been applied
very successfully in numerous experiments to evolve networks in simulation that were
then used with minimal adaptation effort to control real
robots (see
e.g.~\cite{Markelic2007aAn-Evolved,Twickel2011aDeriving,Twickel2012aNeural
,Manoonpong2007cNeural}). Unfortunately, the initial implementation did not
support the evolution of disjunct neuro-modules. As an example, consider the
Aibo\textsuperscript{TM} robot for which a behaviour was evolved
in~\cite{Markelic2007aAn-Evolved}. This system has a left-right symmetry that
an evolutionary algorithm can take advantage of to reduce the search space. In
ISEE, this required a few workarounds. NERD~\cite{Rempis2010aNERD} is a full
evolutionary environment, designed to overcome the limitations of ISEE with a
lot of additional functionality. It allows to define neuro-modules with a rich
repertoire of operators. An interesting feature is that NERD allows the
definition of connection rules, e.g.~to force that a specific sensor has to be
connected to a specific actuator. The software is very powerful, which comes
with the cost that it is not easy to use out of the box. Additionally, NERD is
also unfortunately no longer maintained as the development has stopped.

{\sf NMODE}'s development was initiated to overcome ISEE limitations by adapting
NERD's neuro-modular concept in a minimalistic and easy to use way,
thereby adapting interesting ideas from NEAT\@. The novel
idea that guided {\sf NMODE}'s development is the principled way of augmenting
already evolved networks by new modules, which we see as
one of the biggest problems in the context of evolutionary robotics to achieve
behaviours of higher complexity. If we look at the impressive work by Karl
Sims~\cite{Sims1994aEvolving,Sims1994bEvolving}, then it is difficult to see a
significant progress since these early experiments. Otherwise stated, in the
context of evolutionary robotics, we are still mostly working on the level of
Braitenberg vehicles~\cite{Braitenberg1984aVehicles}, which means that we
mostly work with purely reactive systems. Although the complexity of the morphology
has clearly increased over the decades, the behaviours that these systems show are
still mainly reactive. To the best of the authors knowledge, there are currently
no systems in which a non-trivial morphology has been evolved beyond the realms
of reactive behaviours. This is where the development of {\sf NMODE} is targeted
at; evolving complex behaviours for complex morphologies. This requires a method
to incrementally increase the complexity of the behaviours without loosing
already gained functionality. The method of adding new functionality in this
context is called incremental evolution and is it not a new
concept~\cite{Nolfi2000aEvolutionary}.

The core ideas of {\sf NMODE} can be summarised in the following way. A neural
network is decomposed into neuro-modules. Neural networks are composed of
modules and mutations can only the change the structure within a modules. In
particular, this means that synaptic connection cannot be created between
neurones of different modules. Instead, the interaction between two modules is
controlled by interface neurones (see next Sec.). This allows to incrementally
extend previously evolved modules with new modules in a controlled manor. {\sf
NMODE} is not designed to be biologically plausible, but to allow to evolve
neural networks for non-trivial behaviours of complex systems. The intention of
this work is to introduce {\sf NMODE} and investigate how it works on a well-known
morphology (see Sec.~\ref{sec:hexaboard}) as well as a more complex,
biologically motivated morphology (see Sec.~\ref{sec:svenja}). Evolving
non-reactive behaviours will be investigated in future publications.

This work is organised in the following way. The next section discusses the
neuro-modules and evolutionary algorithm in detail. The source code is freely
available at~\cite{Ghazi-Zahedi2016aNMODE}. The following section presents two
experiments, before this work concludes with a discussion and outlook.

\section{NMODE}
\label{sec:nmode}
Evolutionary algorithms can generally be divided into four functions that each
operate on a population of individuals, namely \emph{evaluation},
\emph{selection}, \emph{reproduction}, and \emph{mutation}. Let $\mathcal{P}$
be a population of individuals, and let $E: \mathcal{P} \mapsto \mathcal{P}$, $S:
\mathcal{P} \mapsto \mathcal{P}$, $R: \mathcal{P} \mapsto \mathcal{P}$, and $M:
\mathcal{P} \mapsto \mathcal{P}$ be the \emph{evaluation}, \emph{selection},
\emph{reproduction}, and \emph{mutation} functions. Then, the evolution from
one generation to the next can be written as
\begin{align}
  \mathcal{P}^{(n+1)} =
M\left(R\left(S\left(E\left(\mathcal{P}^{(n)}\right)\right)\right)\right),
\end{align}
where $n$ refers to the generation index.

{\sf NMODE} is designed to reduce the search space by modularisation of the
neural network structure. Therefore, we will first discuss how the
modularisation is specified, before we present the algorithmic details.

\subsection{Neuro-Modules}
\label{sec:neuro_moduels}
The basic idea of {\sf NMODE} is that the morphology of an embodied agent can
be used to determine how a neural network should be modularised. It is known from,
e.g.~stick insects, that each leg has its own local controller and that the
local leg controller are synchronised by a central nervous system~(see
e.g.~\cite{Cruse199015,Cruse1998aWalkneta,Ekeberg2004aDynamic,Ijspeert2008aCentral,Twickel2011aDeriving}).
Hence, if we have a geometric description of an agent's morphology
(e.g.~Fig.~\ref{fig:experimental_platforms}), which includes the pose of the
segments and the location of the sensors and actuators, we can use that
information to generate a geometric description of a neural network. By that we
mean that neurons will have 3D coordinates that refer to their position in a
global Cartesian coordinate system. Currently, this geometric description must
be defined manually in {\sf NMODE}, but there is no reason why this could not be
automatised in future versions. An example is the {\sf
RoSiML}\footnote{\url{http://keyan.ghazi-zahedi.eu/rosiml}} robot simulation
mark-up language that is used by {\sf YARS}~\cite{Ghazi-Zahedi2016aYARS} to
define an experiment. It contains the pose of each segment as well as the
location of each actuator and sensor.
Gazeboo~\cite{Open-Source-Robotics-Foundation2016aGazebo} uses a similar file
structure that could also be used to automatically generate a neural network
configuration. These files can be parsed for the names and location of segments,
sensors and actuators and also determine symmetries and repetitions.

A module in {\sf NMODE} contains six different types of neurons and a list of
synapses. The next paragraph will describe each of them in detail, but we will
first present the type of neuron model that is currently implemented.
We use the standard additive neuron model, which is given by
the following set of equations:
\begin{align}
  a_i(t+1) & = \theta_i + \sum_j w_{ji} o_j(t)\\
  o_i(t+1) & = \tau(a_i(t)),
\end{align}
where $a_i(t)\in\mathbb{R}$ is the activation of neuron $i\in\mathbb{N}$ at
time $t\in\mathbb{N}$, $o_i\in\mathbb{R}$ is the output of neuron $i$ at time
$t$, $w_{ji}\in\mathbb{R}$ is the synaptic strength between the pre-synaptic
neuron $j\in\mathbb{N}$ and the post-synaptic neuron $i$, and finally,
$\tau:\mathbb{R} \mapsto \mathbb{R}$ is the transfer function. All neurons are
updated synchronously, which means that first the activations $a_i(t+1)$ are
calculated based on the neuron outputs of the previous time step $o_i(t)$.
Once all activations are updated, the neuron outputs follow, i.e., all
outputs $o_i(t+1)$ are calculated based on the updated actuators states
$a_i(t+1)$. As already indicated, we currently work with time discrete neural
networks. It must be noted here that the specification given in the {\sf NMODE}
XML file is generic enough to allow {\sf NMODE} to be extended to work with any
type of neural networks or graphical models. This is why we will sometimes
refer to neurons as nodes and synapses as edges. Currently, we support three
type of transfer functions, namely $\tau\in\{\mathrm{id}, \mathrm{sigm},
\mathrm{tanh}\}$, where $\mathrm{id}(x) = x$ is the identity function,
$\mathrm{sigm}(x) = \nicefrac{1}{1+e^{-x}}$ is the standard sigmoid function,
and finally $\mathrm{tanh}$ is the Hyperbolic Tangent.

\begin{figure}[t]
  \begin{center}
    \includegraphics[width=\textwidth]{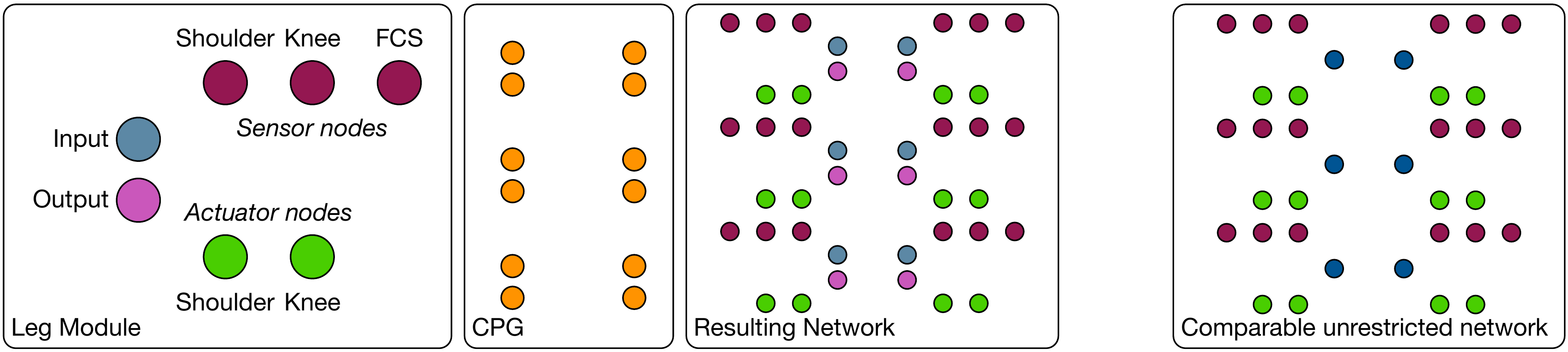}
  \end{center}
  \caption{{\sf Hexaboard} {\sf NMODE} configuration vs.~unrestricted
    configuration. FCS refers to foot contact sensors.}\label{fig:hexaboard_nmode_specification}
\end{figure}

Next, we will discuss the six different types of nodes, which are;
\emph{sensor}, \emph{actuator}, \emph{hidden}, \emph{input}, \emph{output}, and
\emph{connector}. The first three node types are well-known from other neural
network contexts. Sensor nodes receive input from the agent's sensors and are
typically equipped with a linear (id) transfer function. Analogously, actuator
nodes are directly connected to the actuators of the agent, and finally, hidden
nodes are only connected to other nodes within the same module. Actuator and
hidden neurons are usually equipped with one of the two sigmoidal functions
presented above. Two of the new node types are named \emph{input} and
\emph{output}. The role of these nodes is to function as connectors between
modules. Input nodes only allow connections from itself to other neurons
within the module it is assigned to, whereas output nodes only allow connections from
nodes of the corresponding module to itself.
To connect a module with another module that has \emph{input} and \emph{output}
nodes, the connecting module has to use a \emph{connector} node. The
\emph{connector} node will copy the position and properties of the node it is
referring to. In particular, this means that if a connector nodes refers to an
input node, it will automatically become an output node of the connecting
module.

As an example, consider the hexapod used in the first experiment presented below
(see Sec.~\ref{sec:hexaboard}). Its {\sf NMODE} structure is shown in
Figure~\ref{fig:hexaboard_nmode_specification}. Sensor nodes are shown in red,
actuator nodes in green, input nodes in cyan, output nodes in purple, and
finally, connector nodes in orange. In the experiment discussed below, only one
leg module and one CPG module are evolved. By definition, a leg module controls
a leg, which means that requires needs sensor and actuator nodes to access the
leg's state and control its movement. For a simple locomotion movement, the CPG
does not require any sensor or actuator nodes. Instead, the CPG only consists of
connector nodes to interface with the evolved leg modules. Because all legs are
morphologically identical in this example, only one leg module has to be
evolved, which is the used several times for the final controller (see
Fig.~\ref{fig:hexaboard_nmode_specification}).

We demonstrate the parameter reduction that results from this modularisation
based on the hexapod example and with the assumption that the insertion of
hidden units is not allowed. This allows us to calculate the maximal number of
edge for a modularised and non-modularised neural network (see
Fig.~\ref{fig:hexaboard_nmode_specification}), where the latter means that
synapses can connect any two neurons. It must be noted, that sensor nodes can
only have outgoing connections in the current implementation, i.e., they only
function as proxy for the sensor values. Let $n_{s,a,i,o,h}$ be the number of
sensor, actuator, input, output, and hidden nodes. For a fair comparison, we
replace all pairs of input/output nodes by a single hidden node in the
unrestricted configuration, which follows structure that was used
in~\cite{Markelic2007aAn-Evolved,Zahedi2013aLinear}. Then the dimension of the
weight matrices for the two configurations is given by
Tab.~\ref{tab:hexapod_dimension}.

\begin{table}[h!]
  \begin{tabular}{l}
    {\sf NMODE} configuration: Leg Module\\
    $n_s \cdot (n_o + n_a) + n_i \cdot (n_i + n_a) + n_o \cdot (n_i + n_a) +
n_a
      \cdot (n_o + n_a) = 21$ \\
    {\sf NMODE} configuration: CPG Module\\
      $n_i \cdot n_o + n_o \cdot n_i = 72$\\
     {\sf NMODE} configuration: Total = 93\\
    Unrestricted configuration:\\
     $n_s \cdot(n_h + n_a) + n_h \cdot (n_h + n_a) + n_a \cdot (n_h + n_a) =
540$
  \end{tabular}
  \caption{Comparison of search space dimension for the Hexapod example.}
  \label{tab:hexapod_dimension}
\end{table}
For the example of a six-legged walking machine with 2DOF per leg, we see that
the unrestricted configuration has up to 540 free parameters that the
evolutionary algorithm has to take care of, compared to the modularised network
structure, which only requires the evolution of maximally 93 parameters. In
other words, the search space in the unrestricted case is about $5.8$ times
larger compared to the modularised case. The next section discusses the
selection, mutation, and reproduction operators on modules.

\subsection{Selection}
We decided to use the simplest form of selection, which is a rank based
approach. After evaluation, the individuals are ranked based on their fitness
value. The top $n\%$ (user defined) are then selected for reproduction. In the
previously used ENS$^3$
approach~\cite{Dieckmann1995bCoevolution,Ghazi-Zahedi2009aSelf-Regulating},
offspring were assigned to individuals with respect to a Poisson distribution
based on the individual's fitness values. Individuals without offspring after
the reproduction process, were removed from the population. An additional
parameter allowed to save the best $n$ individuals, which might have not been
assigned an offspring despite having a high fitness value. The idea was that,
because of its stochastic nature, this selection method would provide more
diversity in the population. We believe that the speciation method introduced
by~\cite{Risi2012aAn-enhanced} is more valuable in this respect, which is why it
will be included and evaluated in future versions of {\sf NMODE}. We also
believe that cross-over is more useful with respect to ensuring diversity in the
population, which is why a crossover mechanism based on the exchange of modules
is included (see below). In future versions of {\sf NMODE}, we will evaluate a
NEAT-based cross-over technique to allow crossover between two modules instead
of a full exchange.

\subsection{Reproduction}
The task of the reproduction operator is to assign the right amount of
offspring to each individual that was selected after the evaluation. By right
amount we mean that the number of offspring assigned to each individual should
reflect the individual's fitness values. For this purpose, all fitness values
are normalised in the following way:
\begin{align}
  s_i & = \left(\frac{f_i - \min_i\{f_i\}}{\max_{i}\{f_i\} -
\min_i\{f_i\}}\right)^\eta, &
  r_i & = \frac{s_i}{\sum_{i} s_i},
\end{align}
where $i=1,2,\ldots,N$ is the index of the $i$-th individual, and $f_i$ is the
corresponding fitness value, $\eta$ is the elitism parameter, and $r_i$ is the
reproduction factor. A constant value for $s_i$ is used, if $\max\{f_i\} =
\min\{f_i\}$. A high value of $\eta$ leads to a concentration of the
offspring to the individuals with the highest fitness values. Analogously, a
low value of $\eta$ results in an equal distribution of offspring among the
selected individual. The resulting reproduction factors $r_i$ sum up to one,
which means that the normalised $r_i$ can directly be used to distribute the
number of offspring.

One reason for the modularisation of the neural networks and the definition of
the interface nodes is that the modules can easily be exchanged between
individuals. If the cross-over parameter is positive, i.e.~$\zeta>0$, then for
each offspring (which is a copy of its mother) a father is chosen randomly from
the set of parents depending on their normalised reproduction factor $r_i$. This
means that individuals with a high reproduction will be chosen more often as a
mating partners. If two mates are found, then $\zeta$ also determines the
probability with which a module is chosen from the father instead of the mother.
This can be fatal in many cases, because the modules were not co-evolved, and
hence, there is a high probability that e.g.~a leg module that works fine with a
particular CPG will not work at all with another CPG\@. This is accounted for,
because very fit individuals will often mate with themselves (due to their high
fitness value), and hence, good networks will be preserved. On the other hand,
less successful networks will more likely be crossed with more successful
networks, which could increase their survival chances. As briefly mentioned
above, one might think about using NEAT as a method to evolve networks within in
module. This would then allow to cross module-networks instead of completely
switching them. We will investigate this possibility in future versions of {\sf
  NMODE}.

\subsection{Mutation}
{\sf NMODE} is designed particularly to evolve neural networks of arbitrary
structure while restricting the search space in an meaningful way. This means
that the mutations are performed on the level of modules and not on the entire
neural network. Consequently, in the following sections, $\mathcal{E}$ and
$\mathcal{N}$ denote the set of all edges and nodes within a single module.

\paragraph{Synapse insertion}
Because nodes have a three-dimensional position (see above), it makes sense to
insert edges based on the Euclidean distance between nodes. Nodes which are
closer together, will have a higher probability of being connected than those
which are separated by a longer distance. This way, sensors and actuators which
e.g.~read and control the same joint are more likely to be connected early
during the evolutionary process than sensors and actuators which have a larger
distance. We will investigate how a
distance-based edge generation method differs from an uniformly distributed
insertion of edges with respect to the evolutionary success and the generated
behaviours in an experiment discussed below (see
Sec.~\ref{sec:hexaboard}). Both methods are explained in the following
paragraph.

At first, {\sf NMODE} generates a dis-connectivity matrix. This means, that the
matrix $D$ has a positive, non-zero entry at $d_{ij}$ when the synapse from
node $j$ to node $i$ does \emph{not} exist. There are currently two ways in
which the non-zero entries are calculated (see
Alg.~\ref{alg:synapse_insertion}). The first possibility is to set them
uniformly to one. This means that each new connection has the same probability
of being created. This refers to a uniformly distributed insertion of synapses.
The second possibility is to set the values proportionally to the distance
between the two corresponding neurons. This means that neurons which are
closer to each other will have a higher chance of being connected by a synapse.
In both cases, each entry in the dis-connectivity matrix is then multiplied with the
user-defined synapse insertion probability, which means that
the resulting matrix determines with which probability each non-existing
synapse will be created (see Alg.~\ref{alg:synapse_insertion}).
\begin{algorithm}[h!]
  \begin{algorithmic}[1]
    \caption{Synapse insertion}\label{alg:synapse_insertion}
    \Require $p \in [0,1]$ \Comment{Probability to insert a synapse}
    \Require $u() \in [0,1]$ \Comment{Function to draw a uniformly distributed random number}
    \Require $D = (d_{ij}) = 0, \forall i,j = 1,2,\ldots,N$
    \Require $\mathcal{E} \in \mathbb{N}^2$ \Comment{is the set of all existing
nodes,
      which is empty only for the initial generation.}
    \For{$i = 1,2,\ldots,N$}
      \For{$j = 1,2,\ldots,N$}
        \If{$(i,j) \not\in \mathcal{E}$}
          \If{uniform}
            \State{$d_{ij} = 1$}
          \ElsIf{distance}
            \State{$d_{ij} = \nicefrac{1}{\mathrm{dist}(n_i,n_j)}$}
          \EndIf
        \EndIf
      \EndFor
    \EndFor
    \State $D = \frac{D}{\max D}$ \Comment{Normalise $D$ such that the largest
entry is equal to one.}
    \For{$i = 1,2,\ldots,N$}
      \For{$j = 1,2,\ldots,N$}
      \If{$u() < p \cdot d_{ij}$}
      \State{Add synapse $(i,j)$ to $\mathcal{E}$}
      \EndIf{$u() < p_{si}$}
      \EndFor
    \EndFor
  \end{algorithmic}
\end{algorithm}

\paragraph{Synapse deletion}
The user specifies how many synapses can be removed during the mutation of a
module. Hence, if the user specifies $p=1$, than any fraction from $0\%$ to
$100\%$ of the synapses can be deleted, whereas $p=0.5$ means that maximally
$50\%$ of all existing synapses can be deleted. The synapse deletion algorithm
is comparably simple (see Alg.~\ref{alg:synapse_deletion}).
\begin{algorithm}[h!]
  \begin{algorithmic}[1]
    \caption{Synapse deletion}\label{alg:synapse_deletion}
    \Require{$p \in [0,1]$} \Comment{Probability to delete a synapse}
    \Require $u() \in [0,1]$ \Comment{Function to draw a uniformly distributed random number}
    \State{$N = u() \cdot p \cdot |\mathcal{E}|$}
    \For{$i = 1,2,\ldots,N$}
      \State{Delete a synapse randomly drawn from $\mathcal{E}$}
    \EndFor{}
  \end{algorithmic}
\end{algorithm}

One could argue that this form of synapse insertion and deletion are too
invasive, as they can both lead to significant changes of the network
structure. This is intentional for the following reason. {\sf NMODE} was not
designed to be a biologically plausible evolutionary algorithm to understand
how biology has evolved complex systems. {\sf NMODE} was initiated to find
controller for complex systems in cases where other approaches fail. In other
words, the solution is more important than the algorithm that found it.
Therefore, if a local maximum is reached, it is desirable to increase
the stochastic search area to find other maxima. A
good way of doing this is to
significantly alter the structure of the neural network, which is why we allow
such large structural variations.

\paragraph{Synapse modification}
The modification of synapses is straight forward. Each synapse is modified with
a probability specified by the experimenter. The parameter $\delta$ determines
the maximal variation of the synapse and the \emph{max} parameter determines the
limit of the synaptic strength (see Alg.~\ref{alg:synapse_modification}).

\begin{algorithm}[h!]
  \begin{algorithmic}[1]
    \caption{Synapse modification}\label{alg:synapse_modification}
    \Require{$p \in [0,1]$} \Comment{Probability to modify a synapse}
    \Require{$m \in \mathbb{R}^+$} \Comment{Maximal absolute value for the
synaptic strength}
    \Require{$\delta \in \mathbb{R}^+$} \Comment{Maximal absolute change of a
synaptic strength}
    \Require $u() \in [0,1]$ \Comment{Function to draw a uniformly distributed random number}
    \For{$(i,u) \in \mathcal{E}$}
      \If{$u() < p$}
        \State{$w_{ij} = w_{ij} + \delta(2u()-1)$}
        \State{Prune $w_{ij}$, such that $w_{ij} \in [-m,m]$.}
      \EndIf{}
    \EndFor{}
  \end{algorithmic}
\end{algorithm}

\paragraph{Neuron insertion}
In growing a network, it is useful to insert neurons carefully, without massive
structural changes~\cite{Stanley2002aEvolving}. In our previous
implementation~\cite{Ghazi-Zahedi2009aSelf-Regulating,Dieckmann1995bCoevolution},
we used to add a neuron and automatically connect it to a given percentage of
the neural network. This means that every new neuron has a potentially large
impact on the network structure, and hence, its function. We decided to insert
neurons more carefully. Hence, we adopted NEAT's method of inserting
a neuron, which means that with probability $p$ a synapse is chosen which is
split by the new neuron. The synaptic weight of the new incoming synapse is set
to one, whereas the outgoing synaptic strength is equal to the
strength of the original synapse (see Alg.~\ref{alg:neuron_insertion}). The new
neurons position is set to centre of the original synapse.

\begin{algorithm}[h!]
  \begin{algorithmic}[1]
    \caption{Neuron insertion}\label{alg:neuron_insertion}
    \Require{$p \in [0,1]$} \Comment{Probability to add a neuron}
    \Require $u() \in [0,1]$ \Comment{Function to draw a uniformly distributed random number}
    \If{$u() < p$}
      \State{$(i,j)$ is drawn uniformly from $\mathcal{E}$}
      \State{Insert new neuron $k$ in $\mathcal{N}$}
      \State{Remove $(i,j)$ from $\mathcal{E}$}
      \State{Add $(i,k)$ with $w_{ik} = 1$ to $\mathcal{E}$}
      \State{Add $(k,j)$ with $w_{ki} = w_{ij}$ to $\mathcal{E}$}
    \EndIf{}
  \end{algorithmic}
\end{algorithm}

\paragraph{Neuron deletion}
Depending on the number of connections, the deletion of a
neuron can have a strong influence on the function of the recurrent neural
network. For this reason, neurons should be removed with care, which is why in
each mutation at most one neuron is removed (see
Alg.~\ref{alg:neuron_deletion}).

\begin{algorithm}[h!]
  \begin{algorithmic}[1]
    \caption{neuron deletion}\label{alg:neuron_deletion}
    \Require{$p \in [0,1]$} \Comment{Probability to remove a neuron}
    \Require $u() \in [0,1]$ \Comment{Function to draw a uniformly distributed random number}
    \If{$u() < p$}
      \State{Remove a neuron randomly drawn from $\mathcal{N}$}
    \EndIf{}
  \end{algorithmic}
\end{algorithm}

\paragraph{Neuron modification}
Neurons are modified in the same way that synapses are modified. Each neuron
is modified with the specified probability. The bias value is changed with the
limits specified by $\delta$ and cropped within the limits of the specified maximal
value (see Alg.~\ref{alg:neuron_modification})

\begin{algorithm}[h!]
  \begin{algorithmic}[1]
    \caption{neuron modification}\label{alg:neuron_modification}
    \Require{$p \in [0,1]$} \Comment{Probability to modify a neuron}
    \Require{$m \in \mathbb{R}^+$} \Comment{Maximal absolute value for the
bias}
    \Require{$\delta \in \mathbb{R}^+$} \Comment{Maximal absolute change of the
bias}
    \Require $u() \in [0,1]$ \Comment{Function to draw a uniformly distributed random number}
    \For{$n \in \mathcal{N}$}
      \If{$u() < p$}
        \State{$b_{n} = b_{n} + \delta(2u()-1)$}
        \State{Prune $b_{n}$, such that $b_{n} \in [-m,m]$.}
      \EndIf{}
    \EndFor{}
  \end{algorithmic}
\end{algorithm}

\subsection{Evaluation}

{\sf NMODE} is primarily designed to work with {\sf YARS}, although other
simulators and evaluation methods can be easily included too. {\sf YARS} is a
fast mobile robot simulator especially designed for evolutionary robotics. It
uses the bullet physics engine~\cite{Coumans2015aBullet} for the
physics and Ogre3D~\cite{Streeting2006aOgre} for the visualisation. Robots can
be created in blender~\cite{Blender-Foundatation2017ablender} and imported to
{\sf YARS} and blender animation data can be exported from {\sf YARS} allowing
to replay and render the simulation from any camera angle. {\sf YARS} and {\sf
NMODE} are both available from
GitHub~\cite{Ghazi-Zahedi2016aYARS,Ghazi-Zahedi2016aNMODE}. Therefore, the
question of evaluation (except for a brief discussion on fitness functions
below) will not be addressed in this work but in a following publication on the
current version of {\sf YARS}.

This section does not discuss fitness functions in general. There is a vast
amount of literature available on this topic~(see
e.g.~\cite{Holland1992aAdaptation,Nolfi2000aEvolutionary})). Instead, this
section describes how fitness functions are implemented in {\sf NMODE}, because
this is only part that requires programming by the user. Fitness functions are
dynamically loaded during runtime of the {\sf NMODE} executable. The name of the
fitness function given in XML file must correspond to the name of the library.
The class must inherit the {\em Evaluate\/} class and implement the following
functions; {\em void updateController()}, {\em void updateFitnessFunction()},
{\em bool abort()}, {\em void newIndividual()}, {\em void
evaluationCompleted()}. The first function {\em updateController\/} is called
after each step in the simulation, e.g.~after each update of the sensor states.
All sensor values are presented in an one-dimensional array and can be used to
calculate the intermediate fitness. The function {\em abort\/} is then checked
by {\sf NMODE} to see if the evaluation should be terminated early. This
function can be used by the experimenter to indicate that some abort condition
was met, e.g.~if a collision with a wall occurred. If an individual reached the
full evaluation time, the {\em evaluationCompleted\/} function will be called to
allow post-processing. In the examples given below, the
entropy~\cite{Cover2006aElements} over joint angles could be used as part of
the fitness function. This calculation only makes sense at the end of each
evaluation for two reasons. First, the calculation of entropies can be very time
consuming, and hence, should not be performed after every simulation update.
Second, such calculations are most usefull if the full data is available, which
is why the {\em evaluationCompleted\/} function was added. The {\em
newIndividual\/} function is called at the beginning of each evaluation and is
meant for the initialisation and resetting of variables.

This concludes the introduction of {\sf NMODE}. The next section presents two
experiments in which {\sf NMODE} was used to evolve locomotion behaviours in
{\sf YARS}.

\section{Experiments}
\label{sec:experiments}
In the following, we will present two different experiments, namely, {\sf
  Hexaboard} and {\sf Svenja}. The reason why we present these experiments is
that the {\sf Hexaboard} can be considered to be a standard hexapod morphology
with 2DOF per leg (see Fig.~\ref{fig:experimental_platforms}). This is a kind of
benchmark experiment to show how well {\sf NMODE} performs on such tasks. The
second experiment, {\sf Svenja}, is a hexapod morphology that was freely
modelled after an insect. It has four degrees of freedom in each leg and the
three leg pairs are morphologically different (see
Fig.~\ref{fig:experimental_platforms}). {\sf Svenja} is particularly challenging
to control because of the arrangements of the legs and actuators. Contrary to
{\sf Hexaboard}, one cannot use the same network for each leg, as the three leg
pairs will have different tasks, due to their differing morphologies. The front
legs will have to pull, while the rear legs will have to push. The two centre
legs have the same morphology as {\sf Hexaboard}'s legs, and hence, will have a
similar task. Manually programming a behaviour for this morphology turned out to
be very challenging. This experiment was also chosen to demonstrate that {\sf
  NMODE}'s modularisation is well-suited for incremental evolution. Each
leg-pair was evolved individually, with its own CPG\@. For the final
evolutionary phase, the three disjunct networks were simply merged.

\subsection{{\sf Hexaboard}}
\label{sec:hexaboard}
This experiment was chosen because it is a good representation of a standard
experiment in the context of embodied artificial intelligence. The task is to
evolve a neural network for the locomotion of a six-legged walking machine. Most
walking machines are constructed with identical legs, i.e.~each leg has the same
morphology and is attached to the main body in exactly the same way. This
reduces the complexity of the control significantly, because the controller does
not have to distinguish between different leg pairs. The movements of the two
front legs have the same effect on the locomotion behaviour as the movements of
the rear legs. This is in contrast to natural systems, e.g.~cockroaches, for
which the front legs perform a pulling movement, whereas the rear legs can be
said to push. Later in this work, we will evolve a neural network for a hexapod
with a more natural task distribution for its legs (see Sec.~\ref{sec:svenja}).
In this section we will first concentrate on the standard configuration of six
identical legs, which are mounted parallel to each other to a single segment
(see Fig.~\ref{fig:experimental_platforms}).

For such a system, is it natural to assume that each leg should have the same
local leg controller, because each leg should show the same behavioural pattern
in
e.g.~a tripod walking gait. Hence, to reduce the search space, we only need to
evolve a single leg controller and copy it five times for the final controller.
In addition to the leg controller, only the connecting structure (central
pattern generator or CPG) needs to be evolved, which means that the search
space
can be reduced significantly (see Tab.~\ref{tab:hexapod_dimension} an and
Fig.~\ref{fig:hexaboard_nmode_specification}). In what follows, we will first
describe the morphology, followed by the user-defined evolution parameters,
before the results are presented. It must be noted here, that all experiments
were conducted as batch process, i.e., the parameters were set at the beginning
of the experiments, kept constant and the same for all experiments. This is not
how {\sf NMODE} was intended to be used and other parameters might lead to even
better results. The goal here is to show how fast {\sf NMODE} can evolve a good
behaviour for such a standard platform. The goal is not to evolve an optimal
network structure (with respect to size and robustness against disturbances),
which would require alternative growing and pruning phases, modification of
fitness function parameters, etc., and hence, a monitoring of the evolutionary
process. Interactive evolution would not allow to make statistical statements
about the evolutionary progress, which is why we chose to evolve in batch mode.

\subsubsection{Morphology}
\begin{figure}[t]
  \begin{center}
    \includegraphics[width=0.32\textwidth]{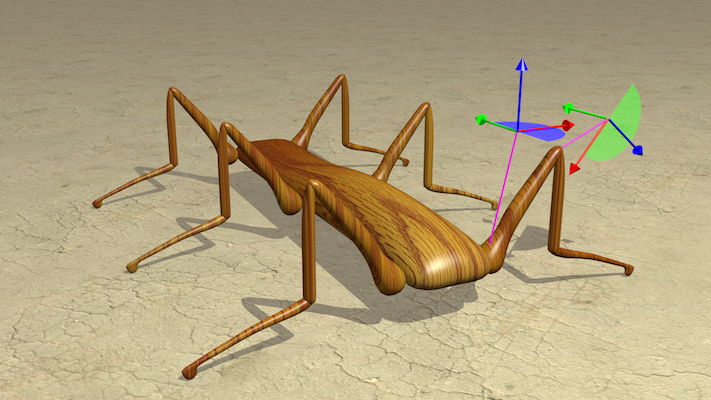}
    \hspace*{1cm}
    \includegraphics[width=0.32\textwidth]{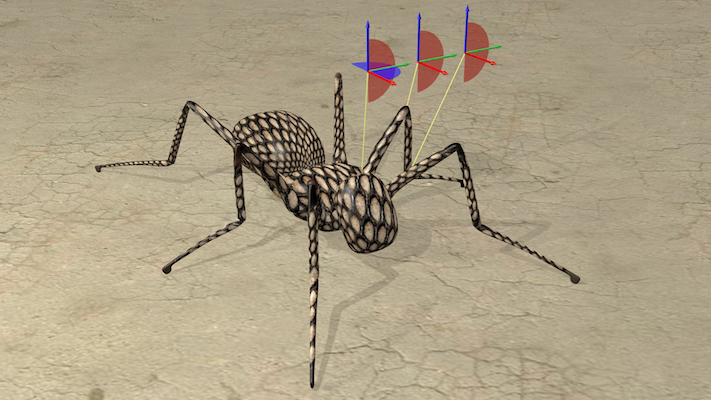}
  \end{center}
  \caption{Experimental platforms. Left: {\sf Hexaboard} is standard hexapod
    with 2DOF per leg. Right: {\sf Svenja} is a hexapod that is loosely
modelled
    after an insect with 4DOF per leg and two light sensors.}
  \label{fig:experimental_platforms}
\end{figure}

We chose {\sf Hexaboard} as our experimental platform for two reasons. First, it
is provided in the {\sf YARS} distribution, which means that the results
presented here can be easily verified, and second, it was successfully used in
previous experiments \cite{Montufar2015aA-Theory}. Overall, the robot was
designed especially for the purpose of learning locomotion, which means that the
body part dimensions, weights, forces, angular range, etc.~were set such that
the system is highly dynamic but has a very low probability of flipping over.
The exact parameters are presented in Table~\ref{tab:hexapod_dimension}.
Figure~\ref{fig:experimental_platforms} visualises the arrangement of the joints
and their rotation axes.

\begin{table}[h]
  \begin{center}
    \begin{tabular}{l|cccl}
      & \multicolumn{3}{c}{Bounding Box (x,y,z)} & Mass\\
      \hline\hline
      Main Body & \unit[0.75]{m} & \unit[4.41]{m} & \unit[0.5]{m}  &
\unit[2.0]{kg} \\
      Femur     & \unit[0.23]{m} & \unit[0.23]{m} & \unit[1.17]{m} &
\unit[0.2]{kg} \\
      Tibia     & \unit[0.1]{m}  & \unit[0.09]{m} & \unit[0.9]{m}  &
\unit[0.15]{kg} \\
      Tarsus    & \unit[0.1]{m}  & \unit[0.08]{m} & \unit[1.04]{m} &
\unit[0.1]{kg}
    \end{tabular}\\
    \begin{tabular}{l|cccc}
      & $\alpha_\mathrm{min} $ & $\alpha_\mathrm{max} $ & $F_\mathrm{max} $ &
$V_\mathrm{max} $\\
      \hline\hline
      Shoulder & \unit[-35]{$^{\circ}$}& \unit[35]{$^{\circ}$}&
      \unitfrac[20.0]{N}{m} & \unitfrac[0.75]{rad}{s}\\
      Knee & \unit[-15]{$^{\circ}$}& \unit[25]{$^{\circ}$}&
      \unitfrac[20.0]{N}{m} & \unitfrac[0.75]{rad}{s}
    \end{tabular}
  \end{center}
  \caption{{\sf Hexaboard's} specification}\label{tab:hexaboard_specification}
\end{table}

All shapes were modelled in blender and the visual appearance of the robot is
equivalent to it's physical form, which means that the shapes that are used for
collision detection etc.~are the same shapes that are also used for the
visualisation. This is also true for {\sf Svenja} (see below).

\subsubsection{{\sf NMODE} parameters}
To evaluate how well {\sf NMODE} evolves a locomotion network for a standard
hexapod platform, we ran each experiment 10 times with the same set of
parameters. From previous experience~\cite{Markelic2007aAn-Evolved}, we decided
to focus primarily on the evolution of the connectivity structure and omitted
the insertion and deletion of neurons in this set of experiments. Hence, the
probability to add or remove a node was set to zero. For the reproduction
parameters, we chose a population size of 100 with a selection pressure of 0.1.
The elitism parameter is set to 10.0, which we determined empirically. Crossover
was set to 0.1. The mutation parameters were set to node modification of 0.01
with a maximal absolute value for the bias of 1.0 and a maximal change of the
bias value to 0.01. Edge modification was set to 0.2, with a maximum of 5.0 and
a maximal step size of 0.5. Edges can be inserted with a probability of 0.05 and
a maximal absolute synaptic strength of 1.0. For the experiment in which the
probability of an edge insertion is dependent on the distance of the nodes, the
minimal distance was set to 0.1, which means that neurons with a distance
smaller than \unit[10]{cm} are set to \unit[10]{cm} (this also applies to
self-connections).

\subsubsection{Fitness Function}
The fitness function is the summed distance in direction of the initial
orientation of {\sf Hexaboard}, which is along the positive $y$-axis. Let
$t=1,2,\ldots,T$ indicate the time steps, then the fitness function is defined
as
\begin{align}
  F & = \sum_{t=1}^T y(t). \label{eq:hexaboard_fitness_function}
\end{align}
The reason for using the summed distance instead of the maximal distance is that
a small progress in distance leads to higher selection pressure,
i.e.~significantly more offspring are assigned to individuals which are slightly
better than their peers.

\subsubsection{Results}
\begin{figure}[t]
  \begin{center}
    \includegraphics[width=0.4\textwidth]{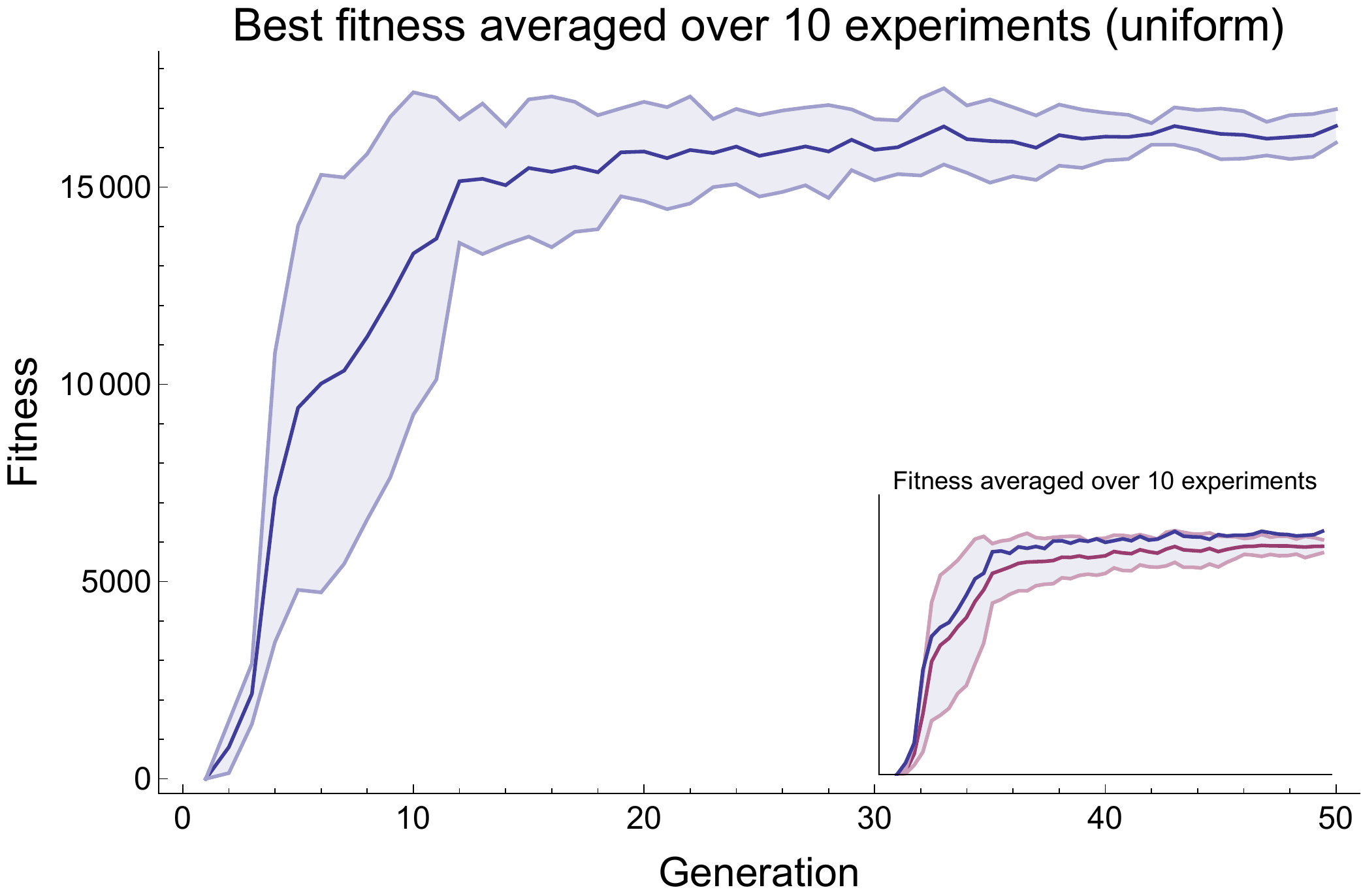}
    \hspace*{0.05\textwidth}
    \includegraphics[width=0.4\textwidth]{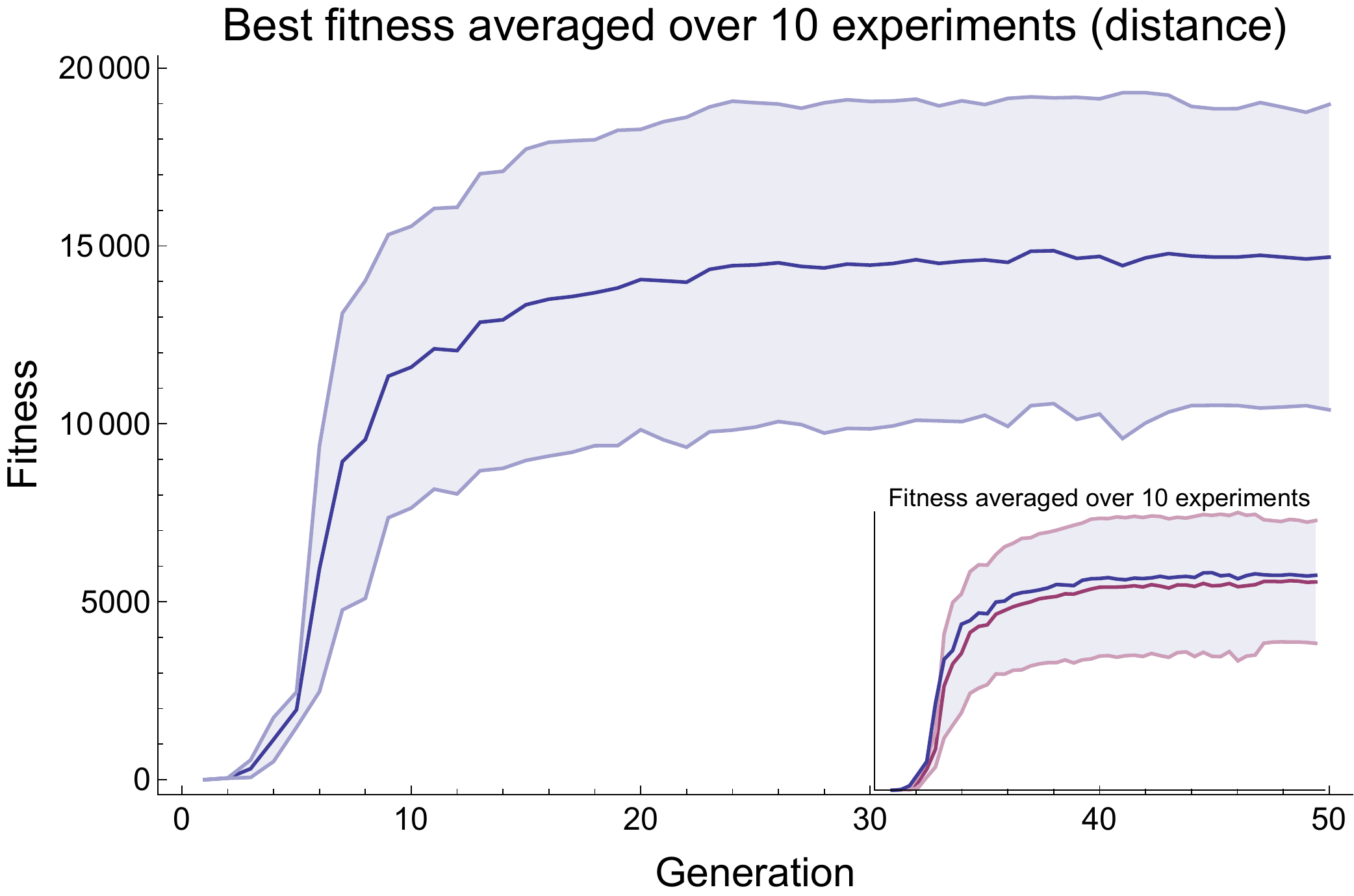}
  \end{center}
  \caption{{\sf Hexaboard} fitness. \emph{Left:} Edges are added uniformly,
    which means that the distance between the connected node is not taken into
    account. \emph{Right:} The probability of inserting and edge decreases with
    the distance between the connected nodes. Both plots show the best fitness
    averaged over 10 experiments, with the corresponding standard deviation.
The
    picture-in-picture plots show the fitness of all selected individuals in
    each generation averaged over 10 experiments. For comparison, the best
    fitness values in each generation is added to the smaller plot as blue
line.}
  \label{fig:fitness_curves}
\end{figure}

\begin{figure}[t]
  \begin{center}
    \includegraphics[width=0.32\textwidth]{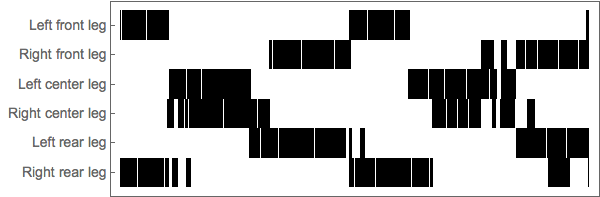}
    \includegraphics[width=0.32\textwidth]{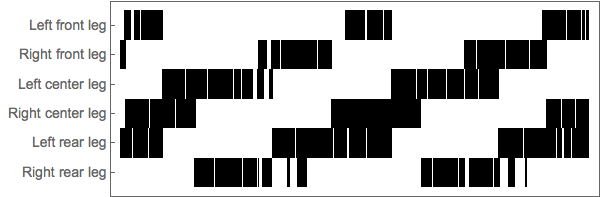}
    \includegraphics[width=0.32\textwidth]{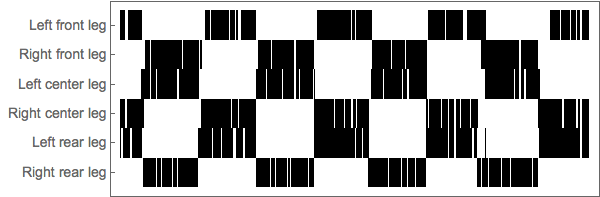}
  \end{center}
  \caption{Examples of evolved walking patterns for {\sf Hexaboard}. \emph{Left
      and centre:} These are two walking patterns that resulted from inserting
    edges dependent on the distance. \emph{Right:} The classical tripod walking
    behaviour that was most dominant when synapses were inserted uniformly. The
    tripod walking pattern was also found in experiments in which the
    distance-based approach was used, but there was a larger variety compared to
    the uniform approach. For the latter, the walking patterns mostly differed
    in the length of the leg stance and swing phases. Videos of the walking
    patterns are available online. In the same order as they are shown above:
    \url{https://youtu.be/a0OtDBHpe1c}, \url{https://youtu.be/UwFKYTIn56w}, \url{https://youtu.be/fZlmPTSmCi0}.}
  \label{fig:hexaboard_walking_pattern} \end{figure}

The plot on the left-hand side of Figure~\ref{fig:fitness_curves} shows the
fitness over generations for a uniform insertion of synapses, while the plot on
the right-hand side of Figure~\ref{fig:fitness_curves} shows the fitness over
generations for distance-based insertion of synapses. The standard deviations
and mean values are taken with respect to all selected individuals of all 10
experimental trials. The smaller plots in each figure show the mean and standard
deviation for only the best individuals over all 10 experiments. Hence, the
large plots show the values for 100 individuals in each generation (10 for each
of the 10 experiments), whereas the smaller plots only show the calculated
values for 10 individuals (the best of each experiment).
Figure~\ref{fig:hexaboard_walking_pattern} shows the resulting walking patterns
for three different evolved individuals.

From the results shown in Figs.~\ref{fig:fitness_curves}
and~\ref{fig:hexaboard_walking_pattern} we can draw the following conclusions.
First, both approaches lead to a good locomotion behaviour already after
approximately 10 generations. To the best of the authors knowledge, no similar
results were published for any other evolutionary algorithm. The second
conclusion is that both approaches, uniform and distance-based insertion of
synapses on average lead to similar results. In both cases, the best walking
behaviours showed a tripod walking behaviour (see
Fig.~\ref{fig:hexaboard_walking_pattern}, left-hand side). Note that the fitness
function (see Eq.~\eqref{eq:hexaboard_fitness_function}) only favours fast
walking but does not reward a specific walking pattern. In the case of the
distance-based approach, we see a higher variance in the fitness over the
different experiments. This is also reflected in the resulting walking
patterns.
Two examples of the best behaviours from the ten experiments with
distance-based insertion of synapses are shown in
Figure~\ref{fig:hexaboard_walking_pattern} (centre and right-hand side). This
leads to the conclusion that if optimality is the highest priority, uniform
insertion of synapses (position-independent) seems to be the approach of choice,
whereas distance-based insertion of synapses should be used if diversity of the
solutions is desired.

\subsection{{\sf Svenja}}
\label{sec:svenja}
\begin{figure}[h]
  \begin{center}
    \begin{minipage}[c]{0.475\textwidth}
      \includegraphics[width=\textwidth]{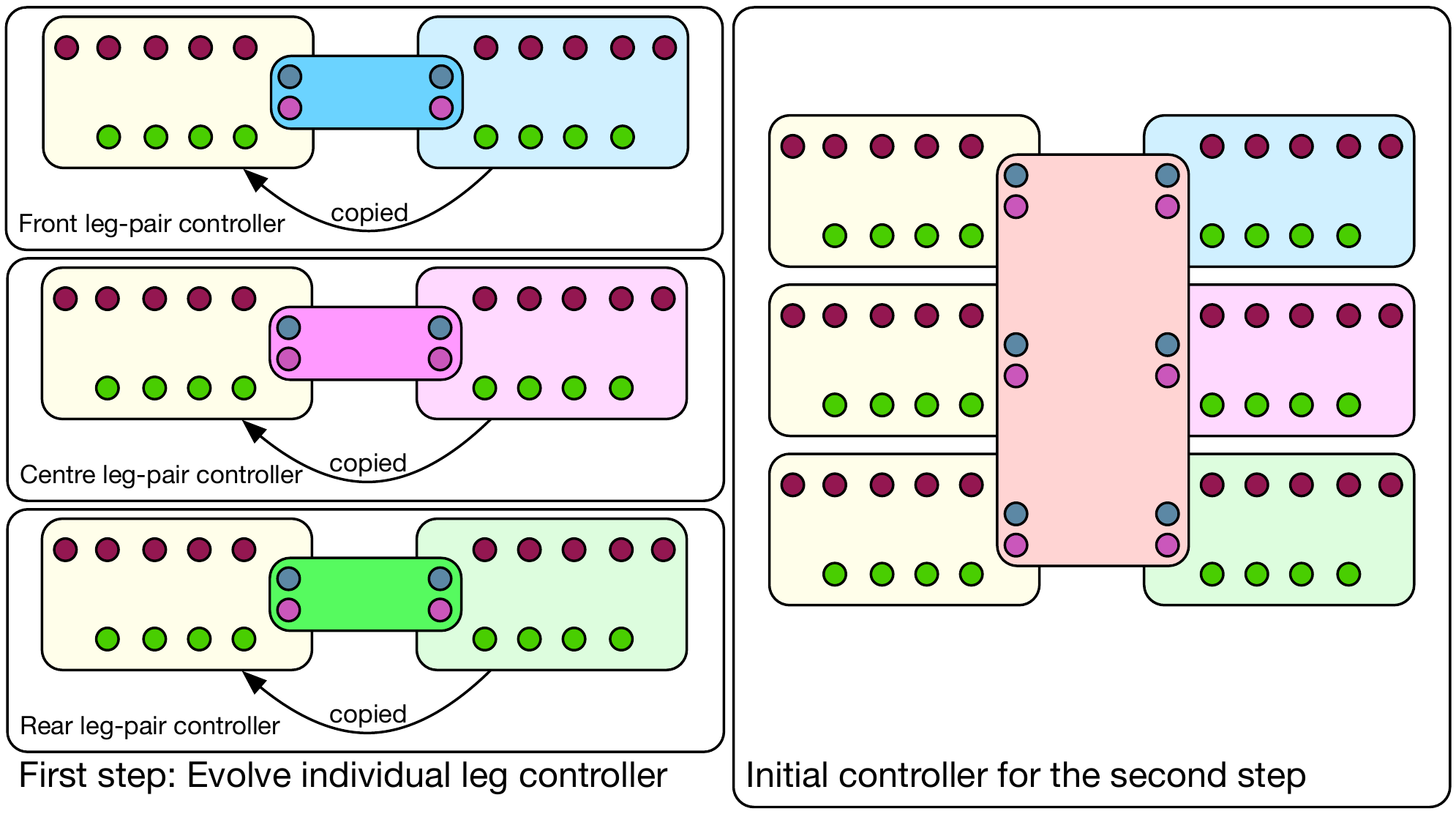}
    \end{minipage}\hfill
    \begin{minipage}[c]{0.475\textwidth}
      \includegraphics[width=0.475\textwidth]{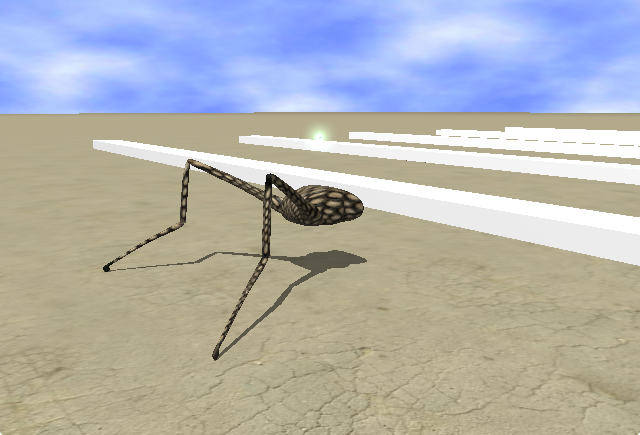}
      \includegraphics[width=0.475\textwidth]{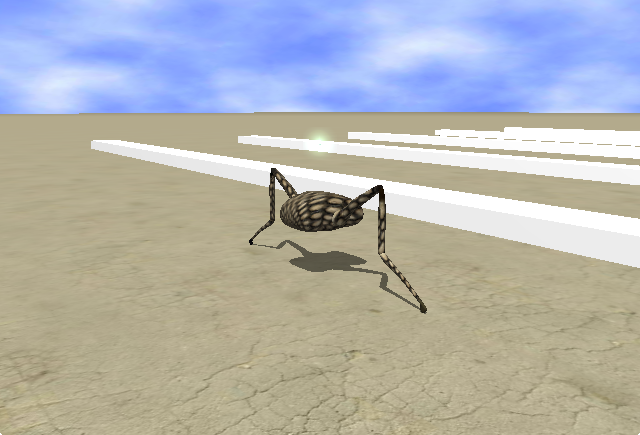}\\
      \includegraphics[width=0.475\textwidth]{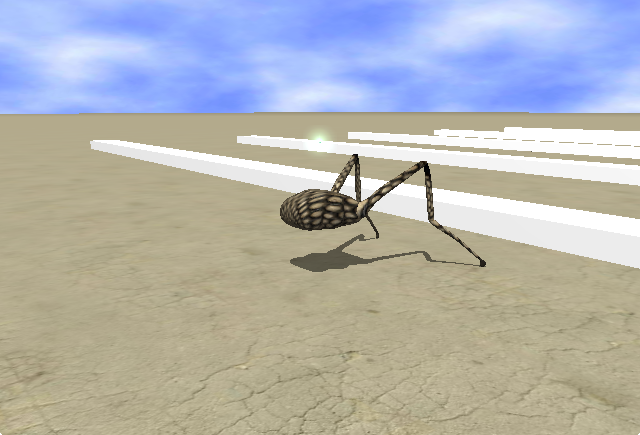}
      \includegraphics[width=0.475\textwidth]{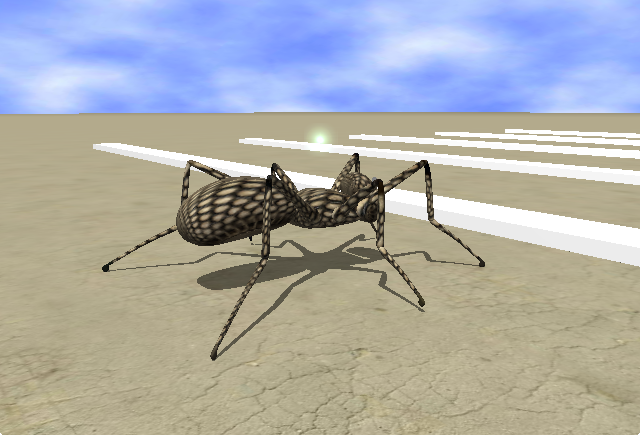}
    \end{minipage}
  \end{center}
  \caption{Incremental evolution of {\sf Svenja}. \emph{Left-hand side:}
    Controller set-up. First, for each leg-pair, individual controller are
    evolved (first image on the left hand side). The evolved leg controller are
    then combined to control the full morphology (second image on the left-hand
    side). Note that the combination of the three controller is straight forward
    (see text for details). \emph{Right-hand side:} The corresponding evaluation
    environments in {\sf YARS}. From left to right and top to bottom, rear legs,
    centre legs, front legs, and full morphology.}
  \label{fig:svenja_evolution}
\end{figure}

{\sf Svenja} was chosen for presentation here to demonstrate how {\sf NMODE} can
be used for incremental evolution of complex systems. {\sf Svenja}'s morphology is
significantly more complex compared to {\sf Hexaboard} for four main reasons.
First, each leg has 4 DOF (instead of 2). Second, the leg-pairs significantly
differ in their shape and attachment to the main body. Third, the main body is
segmented, with a non-even distribution of the weight, in particular, the rear
end of {\sf Svenja} is twice as heavy as the head, which means that the center
of mass is not within the convex hull of the leg's anchor points (see
Fig.~\ref{fig:experimental_platforms} and Tab.~\ref{tab:svenja_specification}).
Fourth, due to the leg morphology and weight distribution, {\sf Svenja} is very
likely to flip over if the legs are not moved properly. To further increase the
difficulty of the task, we included obstacles in the environment (see
Fig.~\ref{fig:svenja_evolution}, right-hand side).

The approach we chose to evolve {\sf Svenja} is incremental in the sense that we
first evolve every leg-pair independently and then merge the resulting networks
to evolve the coordination among the legs. As for {\sf Hexaboard}, evolution was
performed as a batch process, with initially fixed parameters. 

Figure~\ref{fig:svenja_evolution} sketches the incremental approach. On the
left-hand side, the evolution of each leg pair is shown. The leg-pairs were
evolved with a single leg controller, that was mirrored for the other leg, and a
CPG that only consisted of connector nodes to the leg modules (one input and one
output node per leg module). For the final controller (see
Fig.~\ref{fig:svenja_evolution}, left-hand side), the three leg modules were
simply appended in a single XML file. The three disjunct CPGs were appended into
a single module, which was then appended to the list of leg modules. A
small Python script was enough to automate this process, which also could have
been done manually with a simple XML editor. Figure~\ref{fig:svenja_evolution}
(right-hand side), shows the three different evolutionary settings for the first
step of the incremental evolution. The main body is attached to an invisible
rail that only allows movements along the $z$- and $y$-axes (up/down and
forwards/backwards movements). Movements along the $x$-axes and rotations in
general are constrained to zero.

In the following paragraphs, we will first give the specifications of the
morphology, then describe the {\sf NMODE} parameters and fitness function,
before the results are presented.

\subsubsection{Morphology}
{\sf Svenja} has 4DOF of which the first are located in the first joint. This
means that the femur can be rotated around two axes simultaneously (see
Fig.~\ref{fig:experimental_platforms}). The other two DOFs allow rotations
around a single axes and connect the femur with the tibia and the tibia with the
tarsus.
{\sf Svenja}'s main body consists of three
segments. This affects the centre of mass of the entire system. The head and
main segments (two which all legs are attached) have a weight of \unit[10]{kg},
whereas the rear segment has a weight of \unit[20]{kg}. The full specifications
of {\sf Svenja} are summarised in Table~\ref{tab:svenja_specification} (see
Appendix).

\subsubsection{{\sf NMODE} parameters}
During the first phase, three independent evolutionary experiments were
conducted, one for each leg pair. In all runs (repeated five times each), we
used the same {\sf NMODE} parameters, which are set to; life time is 500
iterations, generations are limited to 250, population size is set to 50, elite
pressure is set to 5.0, with a cross over probability of 0.25, node mutation
probability is set to 0.1, with a maximum of 1.0 and maximal step width of 0.2,
nodes are added with with a probability of 0.1 (and a maximal bias values of
1.0) and deleted with with probability of 0.1, edge mutation probability is set
to 0.3, with a maximum of 5.0 and maximal step width of 1.0, edges are added
with with a probability of 0.05 (and a synaptic strength of 1.0) and deleted
with with probability of 0.05.

\subsubsection{Fitness function}
The main part of the fitness is equivalent to the fitness function used for {\sf
Hexaboard}, which means that we used the sum over the main body's
$y$-coordinate in the world coordinate frame. For the evolution of the leg
pairs, an additional punishment was added for changes of the main body's
$z$-coordinate. This means that in principle each evolved leg pair run could
result in different walking heights as long as the main body height changes were
minimal. The fitness function read as:
\begin{align}
  F = \sum_{t=1}^t [y(t) - \gamma(z(t) - z(t-1))],
\end{align}
where $z(0)$ is the initial position of the main body before the
simulation is iterated for the first time.

\subsubsection{Results}
\begin{figure}[t]
  \begin{center}
    \hfill
    \includegraphics[width=0.45\textwidth]{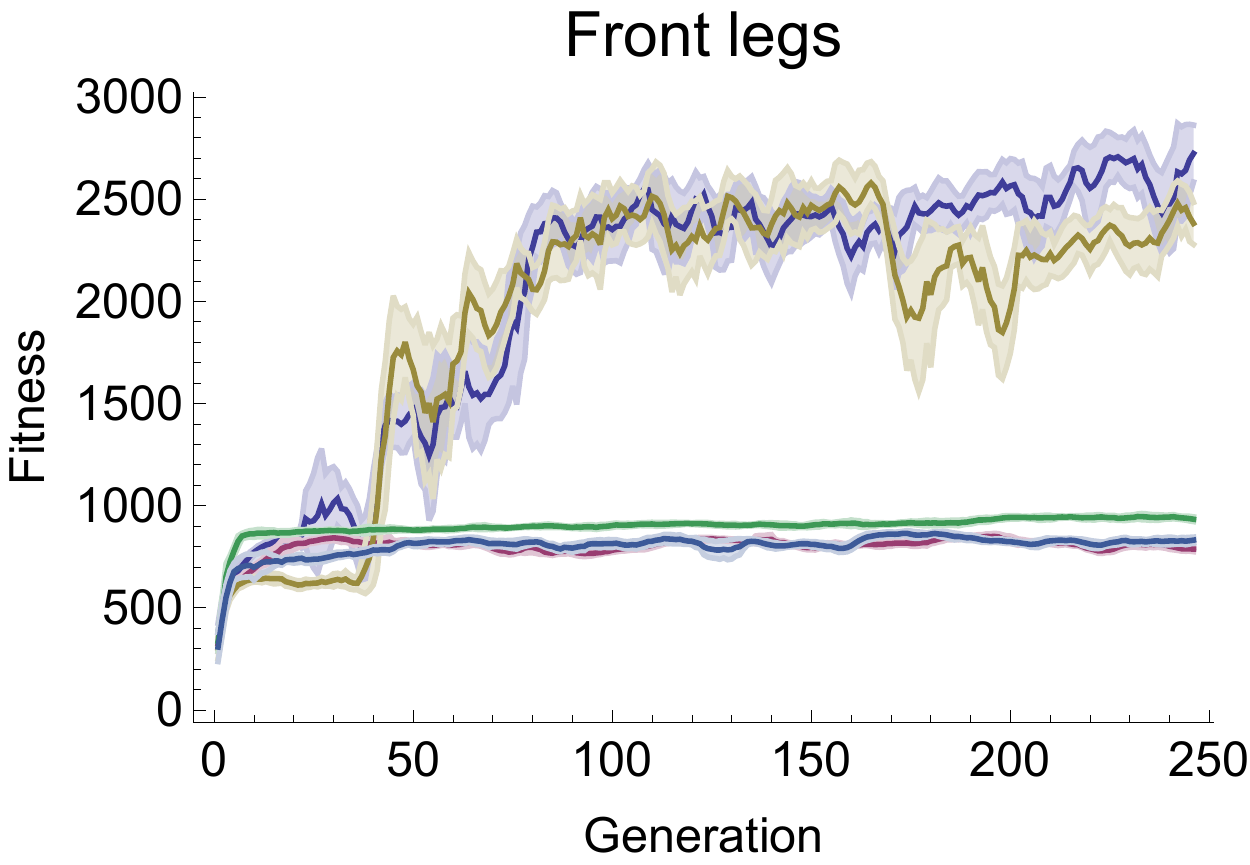}
    \hfill
    \includegraphics[width=0.45\textwidth]{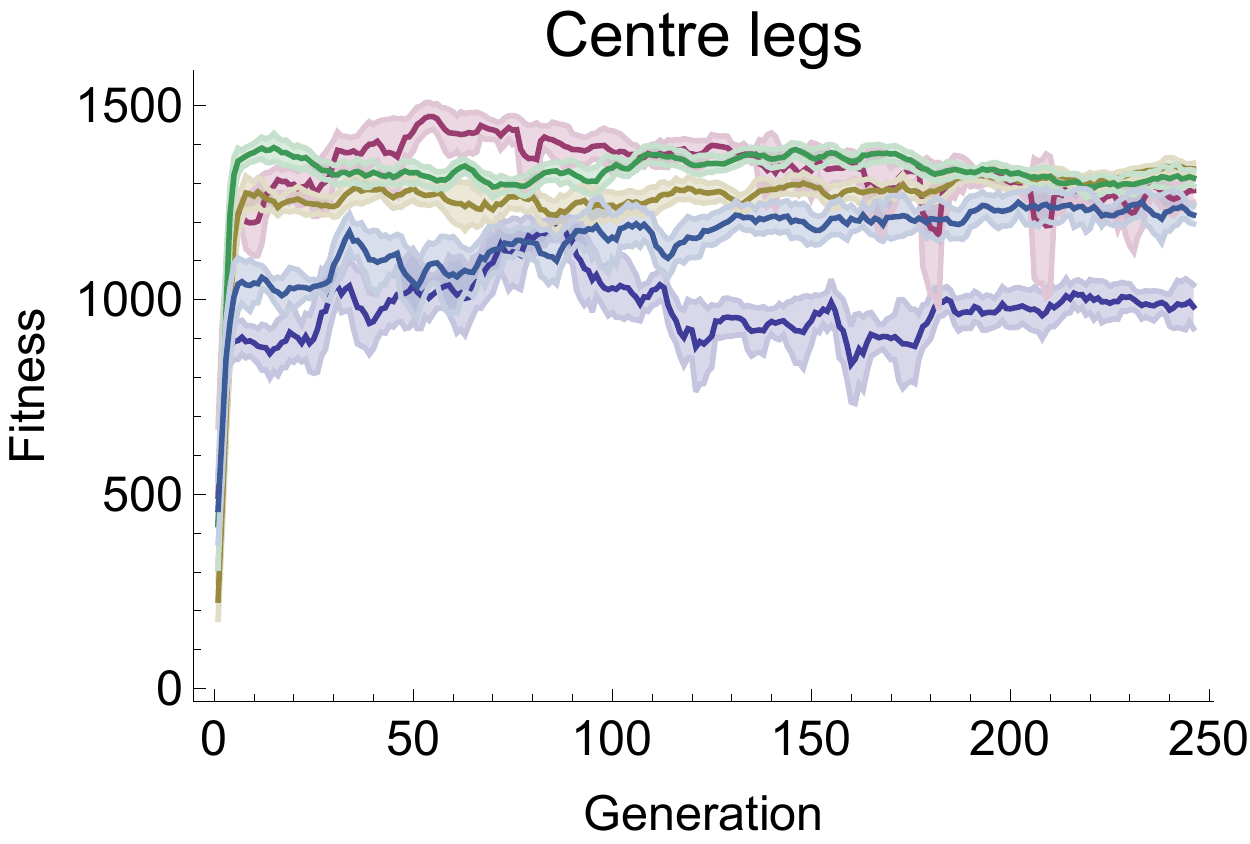}
    \hfill
    \\
    \hfill
    \includegraphics[width=0.45\textwidth]{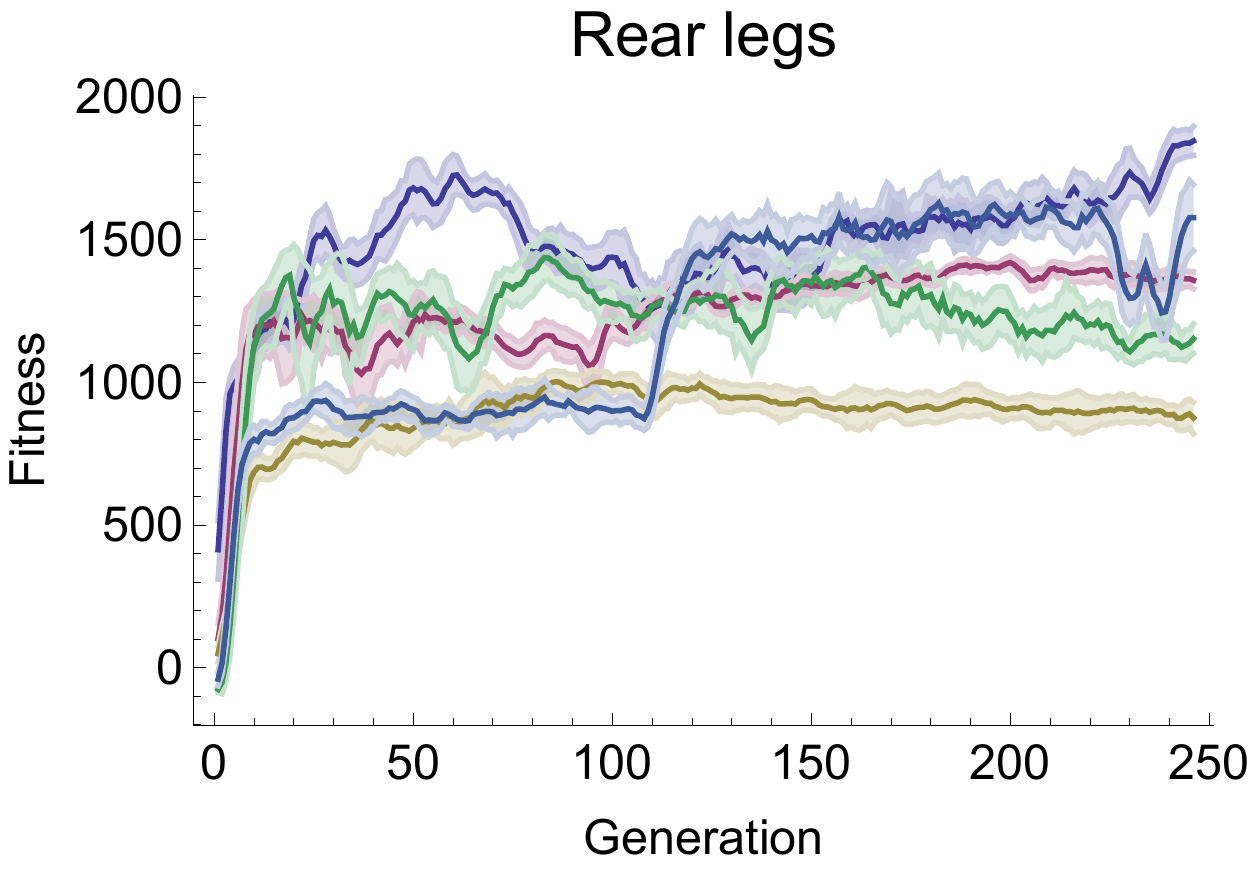}
    \hfill
    \includegraphics[width=0.45\textwidth]{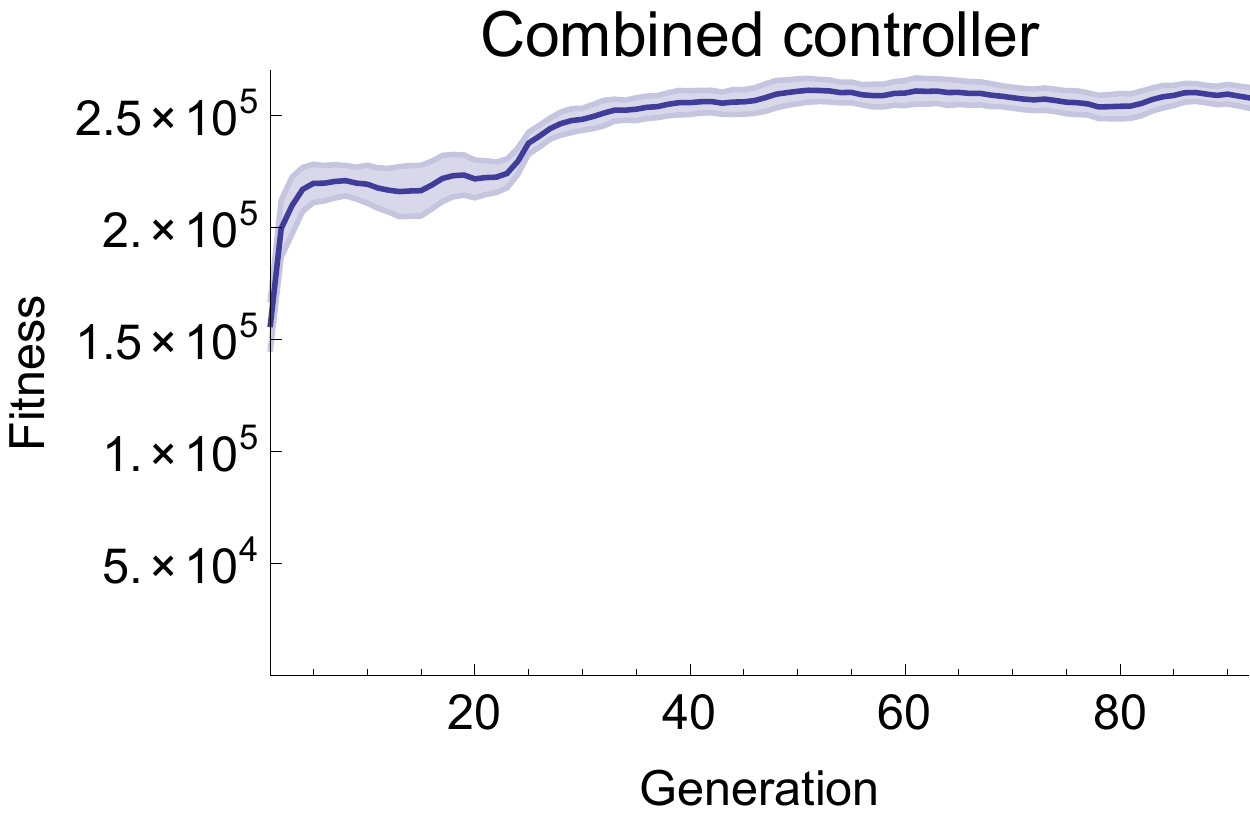}
    \hfill
  \end{center}
  \caption{Fitness of {\sf Svenja} controller over generations. For the final
    controller (lower right-hand side), the best individuals for the five runs
    of each leg pair were chosen. For details, please read the main text.}
  \label{fig:svenja_fitness}
\end{figure}
\begin{figure}[h]
  \begin{center}
   \includegraphics[width=0.49\textwidth]{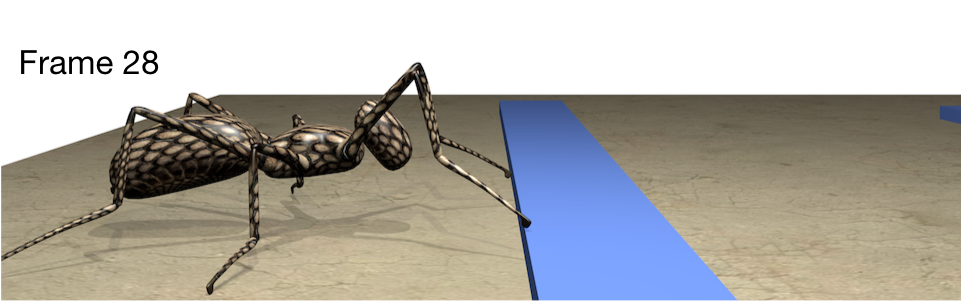}
    \includegraphics[width=0.49\textwidth]{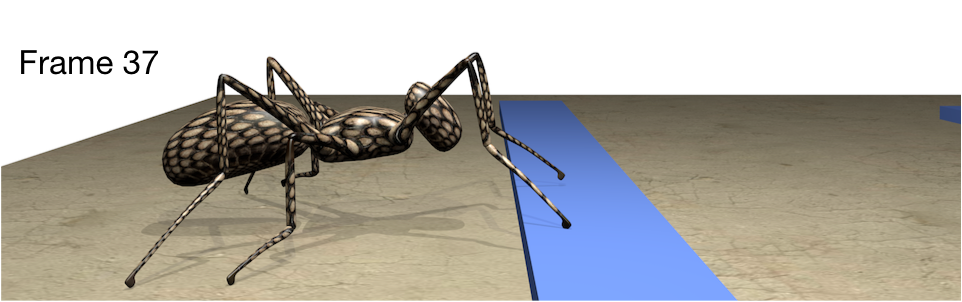}
    \includegraphics[width=0.49\textwidth]{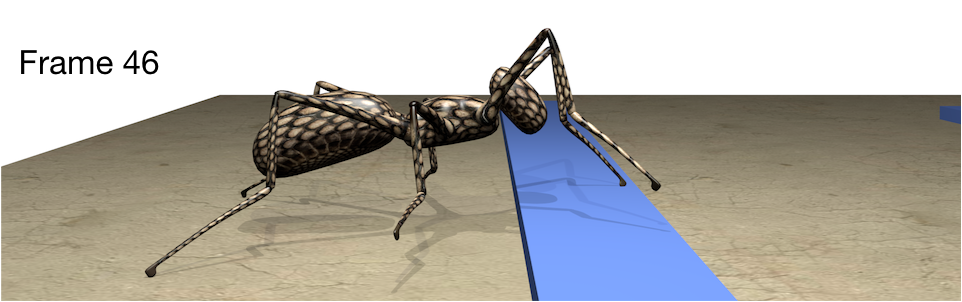}
    \includegraphics[width=0.49\textwidth]{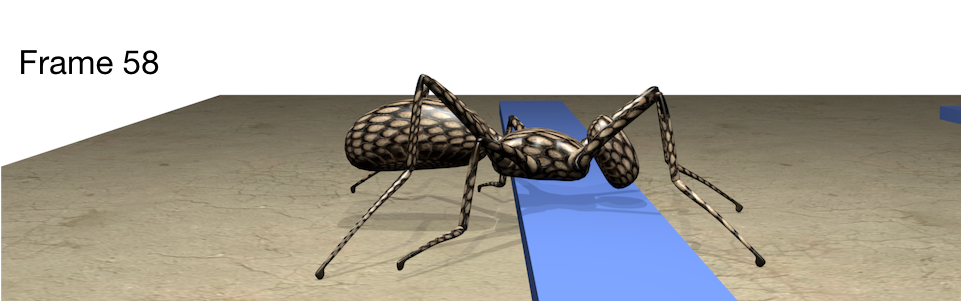}
    \includegraphics[width=0.49\textwidth]{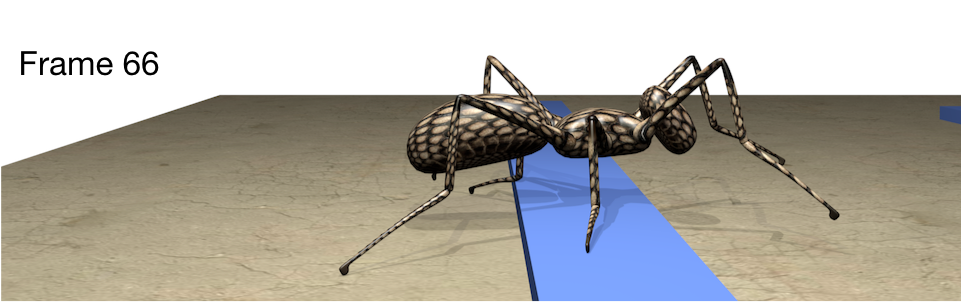}
    \includegraphics[width=0.49\textwidth]{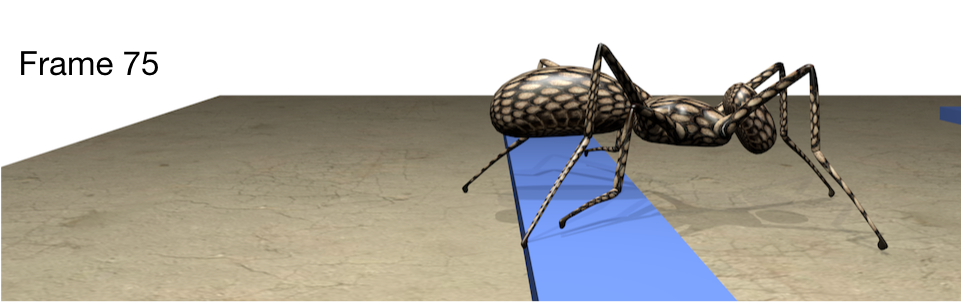}
  \end{center}
  \caption{This figure shows the walking behaviour of {\sf Svenja}. Frames 28,
    37, 46, 58, 66, and 75 are shown here (rendered with blender), which is
    equivalent to approximately \unit[2.5]{s}.}
  \label{fig:svenja_walking}
\end{figure}

The results are captured in Figure~\ref{fig:svenja_fitness} and
Figure~\ref{fig:svenja_walking}. Figure~\ref{fig:svenja_fitness} shows the
fitness over generations for the leg pairs (five runs each) and the evolution
of
the merged leg-pair controller. Figure~\ref{fig:svenja_walking} shows the final
behaviour for a sequence in which {\sf Svenja} climbs over an obstacle.
Videos of all evolved behaviours can also be found
online\footnote{\url{https://www.youtube.com/playlist?list
=PLrIVgT56nVQ7zwLRxUeEEXgxpFhIJk1BY}}.

The first observation is that the front legs seem to be more difficult to evolve
than the other two leg pairs. It took about 100 generations to evolve a good
behaviour compared to approximately 10-20 generations for the centre and rear
leg pairs. The reason is that the way the front legs are attached to the main
body makes it more difficult for them to climb over the obstacles (see
Fig.~\ref{fig:svenja_evolution}, right-hand side). In early generations, each
leg-pair's controller would start with oscillatory movements of low amplitude
and high frequency. For the rear and centre leg-pairs small increases of the
amplitude would already increase the likelihood of climbing the obstacle, while
the front legs would still be blocked by the first obstacle. The reason is that
the oscillatory leg movements of the first leg pair repel the main body more
easily from the obstacle than it is the case for the other leg pairs.

Another observation is that the combined controller very quickly learned to
coordinate the leg pairs. Remember that each leg pair was evolved
independently,
with no restriction of the walking height or walking pattern. Yet,
within less than 10 generations the first coordination occurs. After
approximately 20 generations a good walking behaviour has been found which is
shown in Figure~\ref{fig:svenja_walking}.

In total, the evolution of a good walking behaviour for {\sf Svenja} required
only a few of hours of CPU time. It must be noted again, that this process
was done as a batch process, and hence, the final behaviour is not optimal with
respect to size, robustness, etc. Still, this experiment shows that a behaviour
for a complex morphology such as {\sf Svenja} can be learned very fast with
{\sf
NMODE}.

\section{Discussion}
\label{sec:discussion}
We presented a novel approach to artificial evolution of neural networks and
demonstrated its applicability in two experiments. {\sf NMODE} was not developed
isolated from current research in evolutionary robotics, rather ideas of
different approaches were combined in a minimalistic way. {\sf NMODE} builds
upon the main authors own experience in evolutionary robotics (ISEE), experience
with a collaborator's software framework (NERD) and own experience and
literature research on the currently most dominant approach (HyperNEAT). We
showed that {\sf NMODE} is able to evolve a locomotion behaviour for two
different hexapod experiments in very few generations. To the best of the
authors knowledge, no similar results have been published so far, which shows
the usefulness of {\sf NMODE} in generating interesting behaviours for complex
morphologies.

This said, the development of {\sf NMODE} is not completed yet. On the
short-term, we will evaluate NEAT as a method to evolve the networks within a
module to allow crossover between modules instead of completely switching
modules. In addition, we will also include NEAT's speciation method to protect
innovations. Most importantly, We will have to show that {\sf NMODE} is able to
evolve non-reactive behaviours for complex morphologies, which is the topic of
ongoing experiments.

\section{Acknowledgement}
This work was partly funded by the DFG-SPP 1527 ,,Autonomous Learning''

\bibliographystyle{spmpsci}      


\appendix
\section{{\sf Svenja} specification}
\begin{table}[h]
  \begin{center}
    \begin{tabular}{l|cccl}
      & \multicolumn{3}{c}{Bounding Box (x,y,z)} & Mass\\
      \hline\hline
      Head          & \unit[0.614]{m} & \unit[0.614]{m} & \unit[0.812]{m} &
\unit[10.0]{kg} \\
      Main Body     & \unit[0.976]{m} & \unit[1.107]{m} & \unit[0.552]{m} &
\unit[10.0]{kg} \\
      Rear          & \unit[0.832]{m} & \unit[1.643]{m} & \unit[0.860]{m} &
\unit[20.0]{kg} \\
      Front femur   & \unit[0.612]{m} & \unit[0.945]{m} & \unit[0.977]{m} &
\unit[2.0]{kg}  \\
      Front tibia   & \unit[0.204]{m} & \unit[0.297]{m} & \unit[0.901]{m} &
\unit[1.5]{kg}  \\
      Front tarsus  & \unit[0.611]{m} & \unit[1.026]{m} & \unit[0.444]{m} &
\unit[1.0]{kg}  \\
      Centre femur  & \unit[0.767]{m} & \unit[0.230]{m} & \unit[1.972]{m} &
\unit[2.0]{kg}  \\
      Centre tibia  & \unit[0.103]{m} & \unit[0.094]{m} & \unit[0.909]{m} &
\unit[1.5]{kg} \\
      Centre tarsus & \unit[0.973]{m} & \unit[0.083]{m} & \unit[0.436]{m} &
\unit[1.0]{kg}  \\
      Rear femur    & \unit[0.804]{m} & \unit[1.283]{m} & \unit[1.004]{m} &
\unit[2.0]{kg}  \\
      Rear tibia    & \unit[0.273]{m} & \unit[0.417]{m} & \unit[0.897]{m} &
\unit[1.5]{kg} \\
      Rear tarsus   & \unit[0.729]{m} & \unit[1.230]{m} & \unit[0.446]{m} &
\unit[1.0]{kg}
    \end{tabular}\\
    \begin{tabular}{l|cccc}
      & $\alpha_\mathrm{min} $ & $\alpha_\mathrm{max} $ & $F_\mathrm{max} $ &
$V_\mathrm{max} $\\
      \hline\hline
      Front Ma-Fe 1  & \unit[-5]{$^{\circ}$}   & \unit[20]{$^{\circ}$}   &
\unitfrac[7.5]{N}{m} & \unitfrac[0.5]{rad}{s} \\
      Front Ma-Fe 2  & \unit[-25]{$^{\circ}$}  & \unit[35]{$^{\circ}$}   &
\unitfrac[7.5]{N}{m} & \unitfrac[0.5]{rad}{s} \\
      Front Fe-Ti    & \unit[-10]{$^{\circ}$}  & \unit[10]{$^{\circ}$}   &
\unitfrac[7.5]{N}{m} & \unitfrac[0.5]{rad}{s} \\
      Front Ti-Ta    & \unit[-30]{$^{\circ}$}  & \unit[5]{$^{\circ}$}    &
\unitfrac[7.5]{N}{m} & \unitfrac[0.5]{rad}{s} \\
      Centre Ma-Fe 1 & \unit[-0.5]{$^{\circ}$} & \unit[45]{$^{\circ}$}   &
\unitfrac[7.5]{N}{m} & \unitfrac[0.5]{rad}{s} \\
      Centre Ma-Fe 2 & \unit[-45]{$^{\circ}$}  & \unit[45]{$^{\circ}$}   &
\unitfrac[7.5]{N}{m} & \unitfrac[0.5]{rad}{s} \\
      Centre Fe-Ti   & \unit[-30]{$^{\circ}$}  & \unit[5]{$^{\circ}$}    &
\unitfrac[7.5]{N}{m} & \unitfrac[0.5]{rad}{s} \\
      Centre Ti-Ta   & \unit[-2.5]{$^{\circ}$} & \unit[45]{$^{\circ}$}   &
\unitfrac[7.5]{N}{m} & \unitfrac[0.5]{rad}{s} \\
      Rear Ma-Fe 1   & \unit[-5]{$^{\circ}$}   & \unit[20]{$^{\circ}$}   &
\unitfrac[7.5]{N}{m} & \unitfrac[0.5]{rad}{s} \\
      Rear Ma-Fe 2   & \unit[-30]{$^{\circ}$}  & \unit[20]{$^{\circ}$}   &
\unitfrac[7.5]{N}{m} & \unitfrac[0.5]{rad}{s} \\
      Rear Fe-Ti     & \unit[-5]{$^{\circ}$}   & \unit[27.5]{$^{\circ}$} &
\unitfrac[7.5]{N}{m} & \unitfrac[0.5]{rad}{s} \\
      Rear Ti-Ta     & \unit[-25]{$^{\circ}$}  & \unit[20]{$^{\circ}$}   &
\unitfrac[7.5]{N}{m} & \unitfrac[0.5]{rad}{s}
    \end{tabular}
  \end{center}
  \caption{{\sf Svenja}'s specification. Please note, that the domains for the
    angular position are given for the legs on the right side. Depending on how
    the joint is attached, the values have different meaning, i.e., in which
    direction the leg will move give an specific value. The numbers are
presented
    here to given an impression of the leg's moving
ranges.}\label{tab:svenja_specification}
\end{table}

\end{document}